\documentclass{article}

\usepackage[preprint]{corl_2026}
\usepackage{amsmath,amssymb,amsthm}
\usepackage{graphicx}
\usepackage{booktabs}
\usepackage{multirow}
\usepackage{xcolor}
\usepackage{hyperref}
\usepackage{enumitem}
\usepackage{url}
\usepackage{wrapfig}
\usepackage{caption}
\usepackage{subcaption}
\hypersetup{
    pdfauthor={Jia Zheng, Teli Ma, Yudong Fan, Zifan Wang, Shuo Yang, Junwei Liang},
    pdftitle={MotionWAM: Towards Foundation World Action Models for Real-Time Humanoid Loco-Manipulation},
    pdfsubject={Humanoid loco-manipulation; World Action Model},
    pdfkeywords={Loco-Manipulation, World Action Model, Robotic Manipulation}
  }

\title{MotionWAM: Towards Foundation World Action Models for Real-Time Humanoid Loco-Manipulation}
\author{
  Jia Zheng$^{1,2,\dagger}$, Teli Ma$^{1,2,\dagger}$, Yudong Fan$^1$, Zifan Wang$^{1,2}$, Shuo Yang$^{1,*}$, Junwei Liang$^{2,3,*}$\\
    $^1$ Mondo Robotics \quad $^2$ HKUST (GZ) \quad $^3$ HKUST\\
 $^{\dagger}$ Equal contribution. $^*$ Corresponding author, Co-advising.\\
}

\begin{document}
\maketitle

\vspace{-15pt}
\begin{figure}[h]
\vspace{-10pt}
\centering
\includegraphics[width=1.0\linewidth,trim={0cm 0cm 0cm 0cm}]{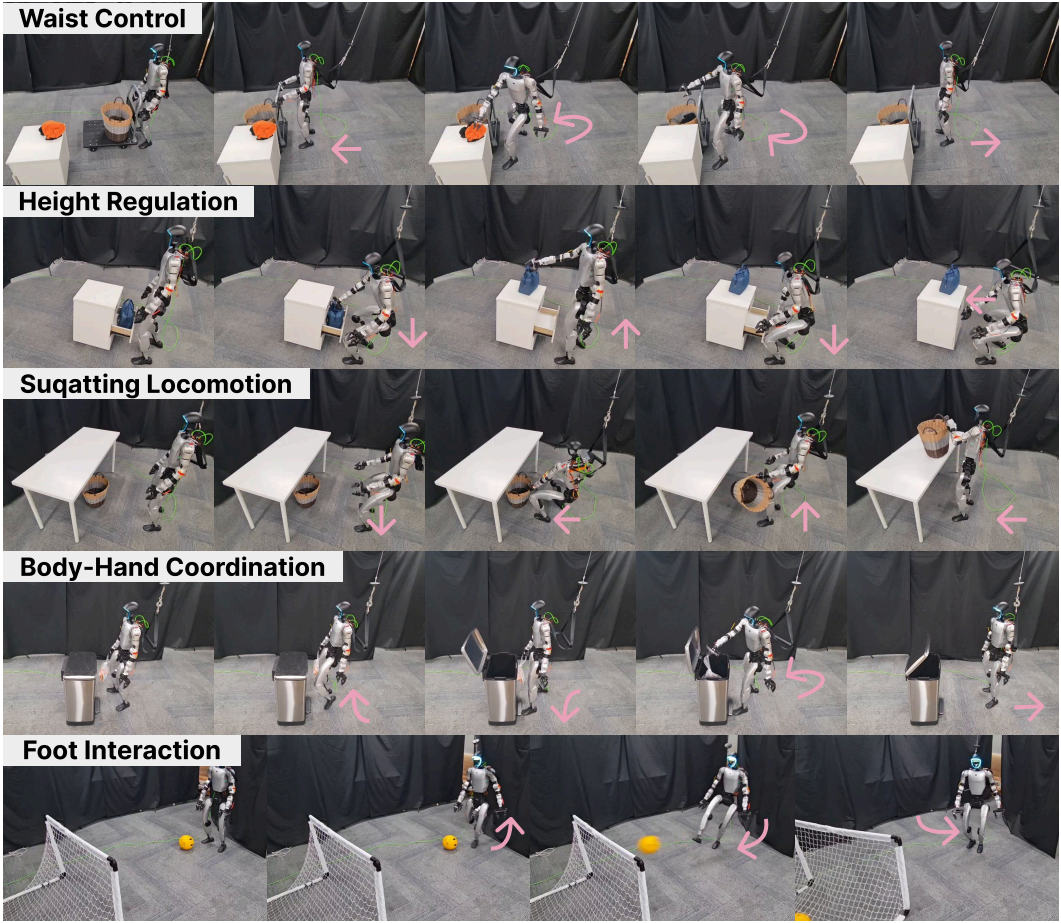}
\caption{\textbf{MotionWAM: A unified WAM for real-time humanoid loco-manipulation.} On a Unitree~G1, MotionWAM produces real-world trajectories spanning waist control, height regulation, squatting locomotion, body-hand coordination, and task-driven foot interaction.}
\label{fig:teaser}
\vspace{-5pt}
\end{figure}

\begin{abstract}
World Action Models (WAMs) couple a video dynamics prior to the policy and have shown encouraging results on tabletop manipulation, but iterative denoising over high-dimensional video-action latents leaves them too slow for real-time humanoid loco-manipulation. The problem is compounded by the dominant hierarchical paradigm, in which a high-level manipulation policy controls only the upper body while a low-level controller tracks coarse base commands---placing upper and lower body in inconsistent action spaces and reducing the legs to balance-preserving locomotion. We present \textbf{MotionWAM}, a real-time WAM that drives autonomous humanoid loco-manipulation from a single egocentric camera by conditioning the policy on the intermediate denoising features of a video world model. MotionWAM replaces the upper--lower split with a \emph{unified motion latent} and predicts whole-body \emph{motion tokens} that jointly cover locomotion, torso motion, height regulation, foot interaction, and hand manipulation in a single action space. A three-stage learning framework progressively adapts the video world model to egocentric visual dynamics and to the target humanoid embodiment. On nine real-world Unitree~G1 tasks, MotionWAM runs in real time, substantially outperforms Vision-Language-Action (VLA) baselines finetuned on the same demonstrations by over 30\% in overall success rate, and executes task-driven foot interaction that decoupled upper-lower policies cannot achieve. Our results suggest that video-pretrained WAMs can be extended from performing tabletop manipulation to coordinated, human-like whole-body humanoid control.
\end{abstract}


\section{Introduction}
\label{sec:intro}

Humanoid loco-manipulation requires robots to move through human-scale environments while coordinating balance, locomotion, reaching, and object interaction.
Prior works have made progress along three largely separate axes: robust whole-body imitation~\cite{ji2024exbody2,he2024omnih2o,liao2025beyondmimic,luo2025sonic}, command-following whole-body controllers~\cite{ben2025homie,ze2025twist2,zhang2025falcon,li2025hold,li2025amo}, and manipulation policies as high-level planners~\cite{wei2026psi_0,jiang2025wholebodyvla,gr00tn16,shi2026egohumanoid}.
Almost all autonomous humanoid loco-manipulation systems combine these axes by pairing a high-level manipulation policy with a low-level locomotion controller~\cite{gr00tn16,wei2026psi_0,shi2026egohumanoid}, 
\begin{wraptable}{r}{0.4\textwidth}
\vspace{-5pt}
  \centering
    \includegraphics[width=0.4\textwidth]{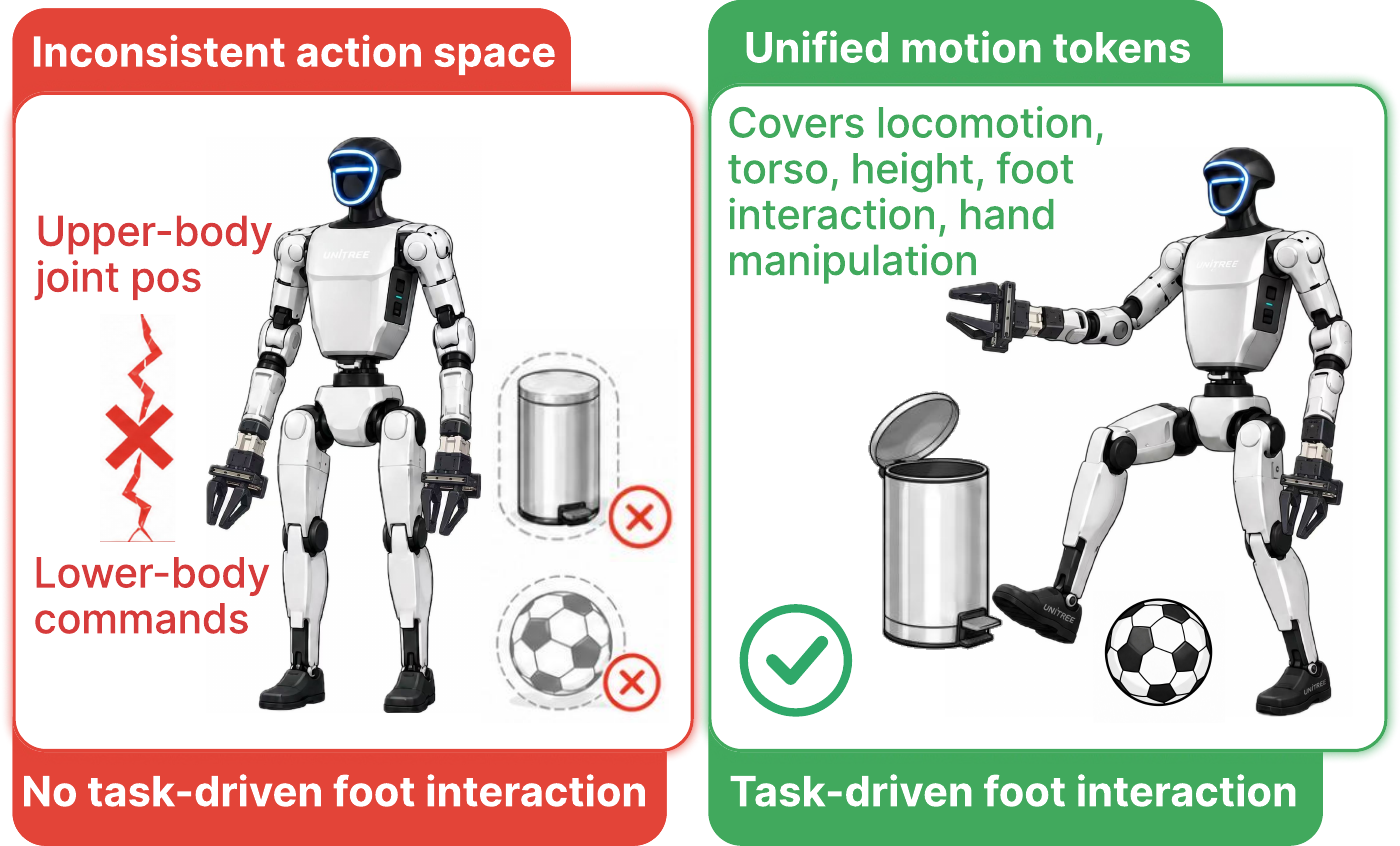}
   \vspace{-10pt}
   \captionsetup{type=figure}
   \caption{\textbf{Decoupled vs.\ unified action spaces.} \emph{Left:} hierarchical pipelines split control into upper-body joint targets and lower-body base commands, restricting the legs to balance preservation. \emph{Right:} MotionWAM predicts \emph{whole-body motion tokens} covering locomotion, torso, height, foot interaction, and hand manipulation, enabling task-driven foot behaviors like pedal stepping and ball kicking.}
   \label{fig:motivation}
    \vspace{-20pt}
\end{wraptable}
giving fine-grained joint targets to the upper body but only coarse base commands (velocity, torso height, orientation) to the lower body. This forces the two halves into inconsistent action spaces and restricts the legs to balance-preserving locomotion, ruling out task-driven foot interaction such as stepping on a pedal or kicking an object, as shown in Fig.~\ref{fig:motivation}.

World Action Models (WAMs) have recently emerged as a promising route to
visuomotor control by inversing dynamics~\cite{hu2024video,pai2025mimic,lingbot} or jointly modeling visual and robot distributions~\cite{cen2025worldvla,li2025unified,bi2025motus,cosmospolicy,ma2026dit4dit,ye2026gigaworld}.
By learning to roll out future visual states for action prediction, WAMs inject a strong dynamics prior into the policy, yielding temporally coherent and physically grounded behavior that purely image-text-pretrained policies struggle to acquire from demonstrations alone. However, iterative denoising over high-dimensional video-action latents is costly, and existing WAMs struggle to reach real-time rates even on tabletop arms. This raises the question: \emph{can WAM's rich dynamics priors be deployed in real time, in a unified latent space, for whole-body humanoid loco-manipulation?}

To resolve these two questions, we present \textbf{MotionWAM}, an end-to-end World Action Model for autonomous humanoid loco-manipulation from a single egocentric camera.
MotionWAM couples a video DiT with a motion DiT through intermediate
denoising features and trains both modalities under a unified flow-matching
objective inspired by DiT4DiT~\cite{ma2026dit4dit}.
Instead of splitting the body into upper-limb joint targets and lower-limb base commands, MotionWAM represents behavior in a \emph{unified motion latent} and predicts \emph{whole-body motion tokens} to drive the humanoid as one integrated system. 
Each token compactly encodes coordinated whole-body motion across locomotion, torso movement, height change, foot placement, and hand manipulation. Crucially, the same representation gives the lower body a task-driven action vocabulary, unlocking foot interaction such as stepping on a pedal or kicking an object.

To learn the visual dynamics priors that egocentric loco-manipulation actually requires, we train MotionWAM with a three-stage learning framework.
\textit{Stage~1: egocentric video pretraining.} We assemble roughly 2{,}136 hours of egocentric human video and humanoid-robot video, and pretrain the video branch alone on future-frame generation. Only the video parameters are updated, shifting the world model from its original distribution toward visual dynamics observed from a first-person viewpoint.
\textit{Stage~2: cross-embodiment action post-training.} To bridge the pretrained world model to our target embodiment, we train on a heterogeneous mixture of Unitree G1 humanoid data spanning different end-effectors and action-annotation formats, jointly updating the video and action networks under a single joint flow-matching objective.
\textit{Stage~3: whole-body fine-tuning.} We collect teleoperated whole-body demonstrations on the target tasks and fine-tune the model from Stages~1--2, switching its action output to unified whole-body motion tokens that drive the humanoid end-to-end.

To rigorously evaluate MotionWAM, we establish a comprehensive suite of nine real-world humanoid loco-manipulation tasks that jointly stress waist control, height regulation, squatting locomotion, task-driven foot interaction, and body--hand coordination. 
Across this suite, MotionWAM runs in real time and substantially outperforms VLA baselines finetuned on the same demonstrations---executing task-driven foot interaction that decoupled upper-lower policies cannot reach, with the legs actively contributing to the task rather than passively preserving balance, and producing whole-body motion that is markedly more coherent and human-like.

Our contributions are:
\begin{itemize}[leftmargin=*]
  \item We propose \textbf{MotionWAM}, a real-time WAM for autonomous humanoid loco-manipulation that drives the policy with the intermediate denoising features of a video world model. To train it, we introduce a three-stage learning framework that progressively adapts the video world model to egocentric visual dynamics and to the target humanoid embodiment.

  \item We replace the upper-lower-body decoupled policies common in
  hierarchical humanoid systems with unified motion latents, letting one policy predict locomotion, torso motion, height regulation, foot interaction, and hand manipulation in a single action space. Crucially, this unlocks task-driven foot control that prior decoupled policies cannot perform.

  \item We present, to our knowledge, the \textbf{first} closed-loop end-to-end WAM-driven policy that performs whole-body humanoid loco-manipulation in real time, including task-driven foot behaviors such as ball kicking and pedal stepping that hierarchical upper--lower baselines cannot produce. 
\end{itemize}

\section{Related Work}
\label{sec:related}
In recent years, robot manipulation has evolved from single arms performing tabletop tasks to humanoid robots performing whole-body tasks. Hierarchical structures have become widely used.

\noindent \textbf{Imitation policies and VLAs for manipulation.}
Imitation policies~\cite{chi2023diffusion,zhao2023learning, ma2024contrastive} and Vision-Language-Action models~\cite{zitkovich2023rt,team2024octo,kim2024openvla,liu2024rdt,bjorck2025gr00t,intelligence2025pi_,yang2025egovla} learn visuomotor manipulation from demonstrations, with VLAs additionally inheriting semantic priors from pretrained vision-language backbones. However, these models rely on a \emph{static} image--text substrate that lacks an intrinsic model of temporal evolution or contact physics, and are typically restricted to arm-and-gripper action spaces on a single embodiment. In contrast, MotionWAM replaces the VLM backbone with a video world model whose intermediate denoising features condition the policy, and extends the action space to whole-body humanoid control.

\noindent \textbf{World Action Models.}
World Action Models (WAMs)~\cite{hu2024video,pai2025mimic,lingbot,cen2025worldvla,li2025unified,bi2025motus,cosmospolicy,ma2026dit4dit,ye2026gigaworld} inject a dynamics prior into the policy by leveraging video generators, often built on internet-scale world models~\cite{bruce2024genie,agarwal2025cosmos,assran2025v, lu2025gwm, nematollahi2025lumos}. Despite their promise, deployments to date remain confined to short-horizon, fixed-base, arm-only tabletop settings, primarily because high-dimensional video-action denoising is too slow for closed-loop control. 
In contrast, MotionWAM targets dynamically balancing humanoids and is the first WAM, to our knowledge, to operate in real time on whole-body loco-manipulation.

\noindent \textbf{Humanoid loco-manipulation.}
Existing humanoid loco-manipulation efforts span whole-body imitation of human motion~\cite{ji2024exbody2, he2024omnih2o, liao2025beyondmimic, luo2025sonic}, command-conditioned whole-body controllers~\cite{ben2025homie, ze2025twist2, zhang2025falcon, li2025hold, li2025amo}, and autonomous policies built on top of such controllers~\cite{wei2026psi_0, jiang2025wholebodyvla, bjorck2025gr00t, shi2026egohumanoid}. The latter, while closest to our setting, typically delegate the legs to a low-level controller driven only by base velocity, height, and orientation, leaving upper- and lower-body action spaces decoupled. In contrast, MotionWAM emits a single whole-body action through unified motion tokens, which lets the legs participate in the task itself rather than only stabilizing the base.

\section{Method}
\label{sec:method}

MotionWAM is a dual-DiT WAM that turns egocentric video dynamics into whole-body humanoid action. We first formalize the \emph{predict-video-dynamics--then-invert} setting and the unified whole-body motion latent it predicts (Sec.~\ref{sec:problem}), then describe the dual-DiT architecture that couples a Video DiT and a Motion DiT (Sec.~\ref{sec:arch}), and finally detail the three-stage training recipe that progressively specializes the model from egocentric video pretraining, through cross-embodiment action post-training, to whole-body teleoperation fine-tuning on Unitree~G1 (Sec.~\ref{sec:training}).

\subsection{Problem Formulation}
\label{sec:problem}

Unlike VLA policies that learn a direct mapping
$\pi_{\theta}(\mathbf{a}_t \mid \mathbf{o}_t, l)$ on humanoid embodiments,
MotionWAM follows a \emph{predict-video-dynamics--then-invert} paradigm and
predicts a unified \emph{whole-body motion latent} that drives the entire
body in a single action space:
\begin{align}
 \mathbf{o}_{t+1} \sim p_{v}(\cdot \mid \mathbf{o}_t, l), \quad
\mathbf{m}_{t}   &\sim p_{a}\!\left(\cdot \mid \mathbf{o}_t, p_t, \mathcal{H}(\mathbf{o}_{t+1}^{\tau_v})\right),
\quad \text{where } \mathbf{o}_{t+1}^{\tau_v} \xrightarrow{\tau_v \to 0} \mathbf{o}_{t+1}.
\end{align}
Here $l$ is the language goal, $\mathbf{o}_t$ is the egocentric observation,
$p_t$ is the proprioceptive state, $\mathbf{o}_{t+1}^{\tau_v}$ is the
intermediate future-frame state at flow step $\tau_v$, and $\mathcal{H}$
extracts hidden states from this generative process. The whole-body motion
latent $\mathbf{m}_t$ jointly covers locomotion, torso motion, height
regulation, foot interaction, and hand manipulation, and is converted to
joint commands by a low-level motion decoder. Training jointly models the
video-action distribution
$p_{va}(\mathbf{o}_{t+1}, \mathbf{m}_t \mid \mathbf{o}_t, p_t, l)$.

\noindent \textbf{Whole-body motion latent.} We instantiate $\mathbf{m}_t=(\mathbf{m}_t^{\text{cont}},\mathbf{k}_t)$ on top of SONIC~\cite{luo2025sonic}, a universal whole-body controller that fuses different motion targets through a single shared latent. The shared latent is bottlenecked by Finite Scalar Quantization~\cite{mentzer2023finite} with $2$ tokens of $32$ levels each, so the SONIC token $\mathbf{k}_t\in\{-1,-\tfrac{15}{16},\ldots,1\}^{64}$ is a $64$-dimensional discretized vector that summarises locomotion, torso, height, and foot-interaction intent. The continuous part $\mathbf{m}_t^{\text{cont}}$ collects the dexterous channels SONIC does not cover---left/right gripper commands or dexterous hand commands---and drives the hands directly.

\subsection{Model Architecture}
\label{sec:arch}

MotionWAM follows the dual-DiT video-motion framework, instantiated for whole-body humanoid loco-manipulation. A Video DiT initialized from Cosmos-Predict2.5-2B~\cite{cosmos25}---a causal spatio-temporal VAE plus a flow-matching diffusion transformer~\cite{peebles2023scalable} conditioned on Cosmos-Reason1 language embeddings~\cite{azzolini2025cosmosreason1}---compresses egocentric conditioning frames $\mathbf{o}_t$ and future frames $\mathbf{o}_{t+1}$ into latents $\mathbf{z}_t^0, \mathbf{z}_{t+1}^0$. Rather than consuming a fully denoised future~\cite{peebles2023scalable}, we install a forward hook on a single transformer block to intercept its activations at a fixed flow timestep $\tau_f$:
\begin{equation}
\mathbf{h}_t^{\tau_f} = \mathcal{H}\!\left[v_{\theta}^{\text{video}}\right]\!\left(\mathbf{z}_{t+1}^{\tau_f}, \tau_f \mid \mathbf{z}_t^0, l\right), \quad \mathbf{z}_{t+1}^{\tau_f}\big|_{\tau_f \to 1} \sim \mathcal{N}(0, I),
\label{eqn:feature}
\end{equation}
where $\mathcal{H}[\cdot]$ reads hidden states of the velocity network $v_{\theta}^{\text{video}}$ in a single forward pass. We fix $\tau_f$ at the pure-noise end of the schedule ($\tau_f\!\approx\!1$), so the Video DiT runs in its \emph{one-shot imagination} regime: given clean conditioning $\mathbf{z}_t^0$ and Gaussian noise for the future, one pass yields activations that encode where the scene is \emph{about to go}, never denoising those frames. This single pass is what keeps MotionWAM real time on a closed-loop humanoid. The Motion DiT then consumes $\mathbf{h}_t^{\tau_f}$ via interleaved self/cross-attention together with the embedded proprioceptive state $p_t$ and noisy whole-body motion-latent tokens, and outputs the velocity field whose integration yields the motion latent $\mathbf{m}_t$. Per-embodiment input/output projectors wrap a shared Motion DiT trunk during multi-embodiment pretraining, and the same trunk is reused at deployment with a single Unitree~G1 projector.

\begin{figure}[t]
\vspace{-10pt}
\centering
\includegraphics[width=1.0\linewidth,trim={0cm 0cm 0cm 0cm}]{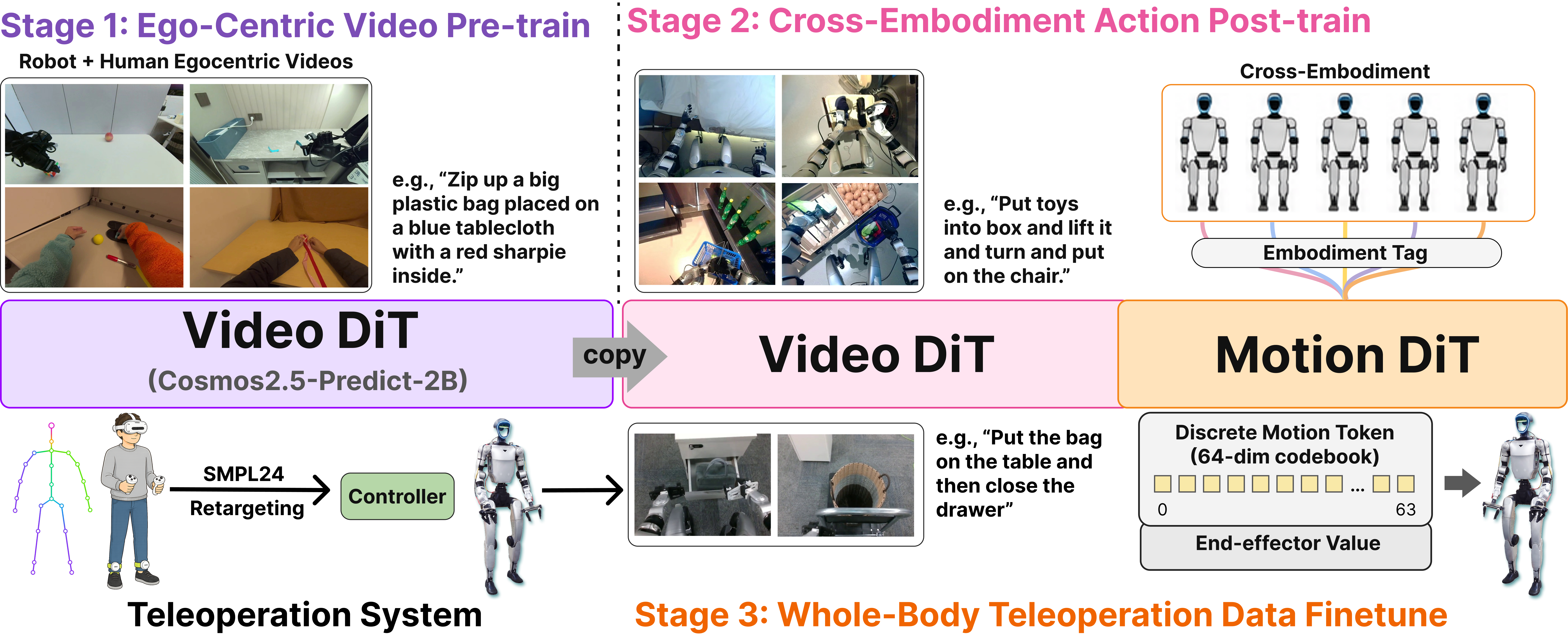}
  \caption{\textbf{Overview of MotionWAM.} A dual-DiT video--motion model trained in three stages. \emph{Stage~1:} the Video DiT is pre-trained alone on egocentric human and humanoid videos. \emph{Stage~2:} the Motion DiT is attached and co-trained across heterogeneous Unitree~G1 datasets via specific embodiment tags, conditioned on Video DiT hidden states to predict discrete motion-token index and continuous end-effector values. \emph{Stage~3:} the full model is finetuned on teleoperated whole-body demonstrations retargeted from SMPL-24 to Unitree~G1.}
  \label{fig:arch}
  \vspace{-15pt}
\end{figure}

\subsection{Training Recipe}
\label{sec:training}

We train MotionWAM with a three-stage learning framework that progressively adapts the video world model to egocentric visual dynamics and to the target humanoid embodiment, following the intuition that the video and motion branches should specialise in turn rather than be optimised jointly from scratch. The VAE and text encoder remain frozen throughout. Stage~1 updates only the Video DiT trunk; Stage~2 jointly updates the Video DiT and the newly attached Motion DiT; Stage~3 fine-tunes the full network end-to-end on the target embodiment.

\noindent \textbf{Flow-matching objectives.} Both the Video DiT and the Motion DiT are trained with flow matching. Reusing the notation of Sec.~\ref{sec:arch}, let $\mathbf{z}_t^0$ and $\mathbf{z}_{t+1}^0$ denote the clean conditioning- and future-frame VAE latents and $\boldsymbol{\epsilon}_v\sim\mathcal{N}(0,I)$ a noise sample; the noise-perturbed future latent at flow time $\tau_v\in[0,1]$ is $\mathbf{z}_{t+1}^{\tau_v}=(1-\tau_v)\,\mathbf{z}_{t+1}^0+\tau_v\,\boldsymbol{\epsilon}_v$, and the Video DiT learns the velocity field $v_\theta^{\text{video}}\!\bigl(\mathbf{z}_{t+1}^{\tau_v},\tau_v\mid \mathbf{z}_t^0,l\bigr)\approx \boldsymbol{\epsilon}_v-\mathbf{z}_{t+1}^0$:
\begin{equation}
\mathcal{L}_{\text{video}} = \mathbb{E}_{\tau_v,\,\mathbf{z}_{t+1}^0,\,\boldsymbol{\epsilon}_v}\!\Bigl[\bigl\lVert v_\theta^{\text{video}}(\mathbf{z}_{t+1}^{\tau_v},\tau_v\mid \mathbf{z}_t^0,l) - (\boldsymbol{\epsilon}_v-\mathbf{z}_{t+1}^0)\bigr\rVert_2^2\Bigr].
\label{eq:fm_video}
\end{equation}
For a clean motion-latent chunk $\mathbf{m}_t^{0}$ and noise $\boldsymbol{\epsilon}_m\sim\mathcal{N}(0,I)$, the Motion DiT predicts the velocity field $v_\phi^{\text{motion}}\!\bigl(\mathbf{m}_t^{\tau_a},\tau_a\mid\mathbf{h}_t^{\tau_f},p_t,e\bigr)$ conditioned on the Video DiT hidden state $\mathbf{h}_t^{\tau_f}$ from Eq.~\eqref{eqn:feature}, the proprioceptive state $p_t$, and an embodiment index $e$:
\begin{equation}
\mathcal{L}_{\text{motion}} = \mathbb{E}_{\tau_a,\,\mathbf{m}_t^{0},\,\boldsymbol{\epsilon}_m}\!\Bigl[\bigl\lVert v_\phi^{\text{motion}}(\mathbf{m}_t^{\tau_a},\tau_a\mid \mathbf{h}_t^{\tau_f},p_t,e) - (\boldsymbol{\epsilon}_m-\mathbf{m}_t^{0})\bigr\rVert_2^2\Bigr].
\label{eq:fm_motion}
\end{equation}

\noindent \textbf{Stage 1 — Pretraining on egocentric human and humanoid video.} The Video DiT is pretrained alone with $\mathcal{L}_{\text{video}}$ on $\sim$2{,}136 hours of egocentric human videos and humanoid-robot videos (the full source-level mixture is given in Appendix~\ref{appendix:data_compose}), shifting the world model from its original distribution toward the visual dynamics observed from a first-person viewpoint. Our key insight is that egocentric visual dynamics, not action diversity, is the bottleneck at this stage: by training the Video DiT alone on cheap, action-free video, the trunk absorbs scale without being throttled by the much smaller pool of action-labelled demonstrations, and produces a robot-centric dynamics prior whose hidden states encode plausible egocentric futures.

\noindent \textbf{Stage 2 — Cross-embodiment action post-training.} We attach the Motion DiT and co-train across heterogeneous Unitree~G1 humanoid data spanning different end-effectors and action-annotation formats, routed through per-embodiment input/output projectors around the shared Motion DiT trunk. To prevent the dynamics prior from being overwritten when the action signal arrives, we retain the video objective as a representation regulariser, giving the joint loss
\begin{equation}
\mathcal{L}_{\text{Stage~2}} = \mathcal{L}_{\text{motion}} + \mathcal{L}_{\text{video}}.
\label{eq:stage2_loss}
\end{equation}

\noindent \textbf{Stage 3 — Finetuning on whole-body teleoperation data.} With the dynamics prior from Stage~1 and the cross-embodiment action grounding from Stage~2 already in place, MotionWAM rapidly adapts to humanoid loco-manipulation from a small amount of teleoperated whole-body demonstrations collected on Unitree~G1 (200 episodes per task across the nine real-world tasks; see Appendix~\ref{appendix:data_compose} and the teleoperation pipeline in Appendix~\ref{appendix:teleop}). We carry over the joint loss in Eqn.~\eqref{eq:stage2_loss} and the shared-trunk + Unitree~G1 projector configuration, so a single network inherits video dynamics, multi-embodiment grounding, and task-specific behavior without any architectural change. The whole-body motion latent decomposes as $\mathbf{m}_t = (\mathbf{m}_t^{\text{cont}},\, \tilde{k}_t)$: a vector of continuous channels $\mathbf{m}_t^{\text{cont}}$ (value of end-effectors) and a single scalar slot $\tilde{k}_t \in \mathbb{R}$ that holds the SONIC~\cite{luo2025sonic} motion-token index $k_t \in \{0,\ldots,K-1\}$ summarising whole-body loco-manipulation trajectories. Rather than introducing a separate categorical head, we let $\tilde{k}_t$ live inside $\mathbf{m}_t$ as a continuous scalar, regress the entire latent under the same flow-matching objective in Eqn.~\eqref{eq:fm_motion}, and recover the discrete index at inference by nearest-neighbor rounding before SONIC decodes the assembled latent into joint commands $\mathbf{a}_t$:
\begin{equation}
\mathbf{m}_t = (\mathbf{m}_t^{\text{cont}}, \tilde{k}_t)
\;\xrightarrow{\;\text{Eq.~\eqref{eq:fm_motion}}\;}\;
\hat{\mathbf{m}}_t = (\hat{\mathbf{m}}_t^{\text{cont}}, \hat{\tilde{k}}_t)
\;\xrightarrow{\;\hat{k}_t = \mathrm{round}(\hat{\tilde{k}}_t)\;}\;
(\hat{\mathbf{m}}_t^{\text{cont}}, \hat{k}_t)
\;\xrightarrow{\;\text{SONIC}\;}\;
\mathbf{a}_t.
\label{eq:motion_token_pipeline}
\end{equation}

\section{Experiments}
\label{sec:exp}

We design our real-robot evaluation around three questions:
(Q1) does coupling a video world model into the policy yield a tangible advantage over strong VLA baselines on whole-body humanoid loco-manipulation?
(Q2) is each of the three training stages necessary, and what does each contribute?
(Q3) can WAM runs fast enough for closed-loop humanoid control compared to VLA baselines?

\subsection{Experimental Setup}
\label{sec:setup}

\noindent \textbf{Hardware platform.}
We conduct all real-world experiments on a Unitree~G1 humanoid with dual ALOHA2 grippers, observing the scene through a head-mounted
\begin{wrapfigure}{r}{0.6\textwidth}
\vspace{-1.0\baselineskip}
\centering
\includegraphics[width=0.6\textwidth]{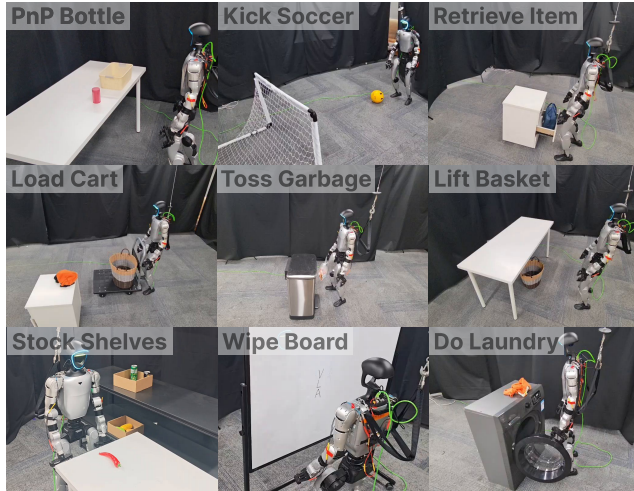}
\caption{\textbf{Real-world task suite.} We design nine whole-body loco-manipulation tasks on the Unitree~G1, each requiring active leg and torso involvement beyond balance preservation. Per-task language prompts are provided in Appendix~\ref{appendix:real_prompt}.}
  \label{fig:exp}
\vspace{-1.3\baselineskip}
\end{wrapfigure}
Intel RealSense D435i RGB camera.
Whole-body teleoperation demonstrations are collected via a PICO~VR three-point tracking setup retargeted to the robot through SMPL, 
and at deployment the policy outputs are tracked by the SONIC~\cite{luo2025sonic} whole-body controller. MotionWAM and all baselines run as WebSocket policy servers on a single NVIDIA RTX~4090 workstation and are queried in closed loop by the on-board controller. Full hardware specifications and the VR-based teleoperation pipeline used to collect Stage~3 demonstrations are described in Appendix~\ref{appendix:teleop}.

\noindent \textbf{Task design.}
As shown in Fig.~\ref{fig:exp}, we evaluate on a suite of nine real-world whole-body loco-manipulation tasks that jointly stress the five core capabilities highlighted in Fig.~\ref{fig:teaser}: \emph{waist control}, \emph{height regulation}, \emph{squatting locomotion}, \emph{task-driven foot interaction}, and \emph{body--hand coordination}. 
The suite is designed so that no single task can be solved by upper-body manipulation alone; each one forces the legs and torso to actively contribute, exposing behaviors that decoupled upper--lower policies cannot express.

Each method is tested for 20 trials per task, and we report success as the percentage of successful trials in Fig.~\ref{fig:comparison}. Per-task language prompts are listed in Appendix~\ref{appendix:real_prompt}, and representative MotionWAM rollouts for every task are shown in Fig.~\ref{fig:full_demos}.

\noindent \textbf{Baselines.}
We compare MotionWAM against five baselines: \textbf{Diffusion Policy}~\cite{chi2023diffusion} and \textbf{ACT}~\cite{zhao2023learning} as representative non-VLA visuomotor policies; $\mathbf{\pi_{0.5}}$~\cite{intelligence2025pi_} and \textbf{GR00T-N1.7}~\cite{bjorck2025gr00t} as state-of-the-art generalist VLAs; and \textbf{Qwen3DiT}, a custom parameter-matched ablation that pairs the Qwen3-VL~\cite{bai2025qwen3} 2B backbone with our Motion DiT to isolate the contribution of the video world model prior over a static VLM prior at matched capacity. All baselines consume the same egocentric RGB observations, language goals, and proprioceptive states as MotionWAM, and emit unified motion latents in the same action space. Per-baseline descriptions and training recipes are reported in Appendix~\ref{appendix:baseline_train}, and MotionWAM training configurations in Appendix~\ref{appendix:model_train}.

\subsection{Comparison with the State-of-the-Art}
\label{sec:main}
\begin{figure}[]
\vspace{-10pt}
\centering
\includegraphics[width=1.0\linewidth,trim={0cm 0cm 0cm 0cm}]{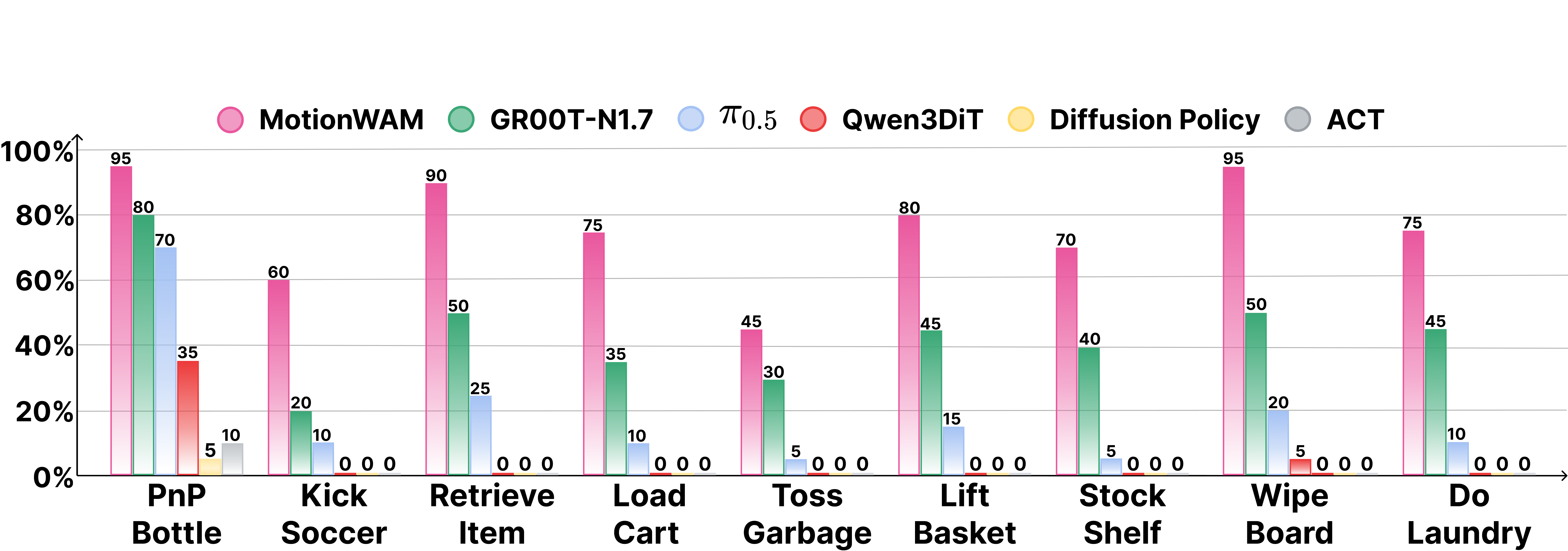}
  \caption{\textbf{Comparison with the state-of-the-art VLAs on nine real-world whole-body loco-manipulation tasks.} We report per-task success rate (\%) over 20 trials per task on the Unitree~G1. All methods are finetuned on the same Stage~3 demonstrations.}
  \label{fig:comparison}
  \vspace{-15pt}
\end{figure}

All policies are finetuned on the same Stage~3 dataset and emit actions through the same SONIC interface, so the sole source of variation is whether the policy is conditioned on intermediate denoising features of a video world model (MotionWAM) or on VLM-style image--text features (GR00T-N1.7, $\pi_{0.5}$, Qwen3DiT).
Fig.~\ref{fig:comparison} reports per-task and average success rates. MotionWAM wins on every task and lifts the overall success rate from $43.9\%$ (the strongest baseline, GR00T-N1.7) to $76.1\%$, an over $32\%$ absolute gain. The gap is largest on tasks demanding whole-body coordination beyond the upper limbs---\texttt{Kick Soccer} ($+40\%$), \texttt{Load Cart} ($+40\%$), \texttt{Retrieve Item} ($+40\%$), \texttt{Wipe Board} ($+45\%$), and \texttt{Do Laundry} ($+30\%$)---where the unified motion latent gives MotionWAM access to task-driven foot and torso behaviors that a decoupled upper--lower interface cannot express. Meanwhile, the VLM-only baseline (Qwen3DiT) collapses to near-zero success on every locomotion-heavy task and even $\pi_{0.5}$ stays under $20\%$ overall, indicating that strong semantic priors alone do not transfer to the closed-loop physics humanoid loco-manipulation demands. Coupling the policy to a video world model is what closes that gap.

\subsection{Ablation on the Three-Stage Training Framework}
\label{sec:ablation}

We disentangle the contribution of \emph{egocentric video pretraining} (Stage~1) from \emph{cross-embodiment action pretraining} (Stage~2), and show that both are necessary for the final real-robot performance. We disable one stage at a time while keeping Stage~3 fixed, and re-run the full real-robot evaluation on a representative subset of the suite: \texttt{Lift Basket}, \texttt{Retrieve Item}, \texttt{Load Cart}, \texttt{Toss Garbage}, \texttt{Kick Soccer}.

\begin{table}[]
  \centering
  \caption{\textbf{Ablation of the three-stage training framework.} Real-robot success rate (\%) over 20 trials per task on five representative whole-body loco-manipulation tasks. Stage~3 (real-robot fine-tuning) is enabled in all variants.}
  \label{tab:ablation}
  \resizebox{0.8\textwidth}{!}{\renewcommand{\arraystretch}{1.0}{
  \begin{tabular}{ccccccccc}
    \toprule
    \multicolumn{1}{c}{Variant} & \multicolumn{1}{c}{Stage~1} & \multicolumn{1}{c}{Stage~2} & \multicolumn{1}{c}{\shortstack{\texttt{Lift}\\\texttt{Basket}}} & \multicolumn{1}{c}{\shortstack{\texttt{Retrieve}\\\texttt{Item}}} & \multicolumn{1}{c}{\shortstack{\texttt{Load}\\\texttt{Cart}}} & \multicolumn{1}{c}{\shortstack{\texttt{Toss}\\\texttt{Garbage}}} & \multicolumn{1}{c}{\shortstack{\texttt{Kick}\\\texttt{Soccer}}} & \multicolumn{1}{c}{\textbf{Avg.}} \\
    \midrule
    w/o Stage~2   & \checkmark & --         & 65 & 45 & 30 & 30 & 40 & 42.0 \\
    w/o Stage~1   & --         & \checkmark & 70 & 75 & 60 & 35 & 55 & 59.0 \\
    \textbf{Full} & \checkmark & \checkmark & \textbf{80} & \textbf{90} & \textbf{75} & \textbf{45} & \textbf{60} & \textbf{70.0} \\
    \bottomrule
  \end{tabular}
  }}
  \vspace{-15pt}
\end{table}

As shown in Table~\ref{tab:ablation}, removing Stage~1 and Stage~2 leads to $11\%$ and $28\%$ absolute performance drop respectively. \emph{Without Stage~1}, the Video DiT enters Stage~2 with a generic, non-egocentric dynamics prior; the resulting hidden states still carry useful semantics, but the predicted motion latents become visibly less accurate. \emph{Without Stage~2}, the Motion DiT is attached directly to the Stage~1 trunk and trained only on the small target-task dataset; with no cross-embodiment grounding to anchor the action signal, performance collapses on every task. The two stages thus play complementary roles---Stage~1 supplies an egocentric \emph{visual-dynamics} prior, Stage~2 grounds that prior into the action space across embodiments. This validates the central design choice of training the video and motion branches in turn rather than jointly from scratch.

\subsection{Real-Time Inference Frequency}
\label{sec:speed}

We measure closed-loop deployment frequency on an NVIDIA~A100 to verify that
conditioning on a \emph{single forward pass} of the Video DiT (Eqn.~\eqref{eqn:feature}) keeps MotionWAM real-time. 
\begin{wraptable}{r}{0.5\textwidth}
\vspace{-1.0\baselineskip}
\centering
\resizebox{0.5\textwidth}{!}{\renewcommand{\arraystretch}{1.0}{
\begin{tabular}{lcc}
  \toprule
  Model & Trainable Params & Frequency \\
  \midrule
  GR00T-N1.7~\cite{bjorck2025gr00t} & 1.6B          & 6.5~Hz \\
  Qwen3DiT                          & 2.3B          & 9.0~Hz  \\
  Cosmos Policy~\cite{cosmospolicy} & 2.0B          & 0.7~Hz   \\
  \textbf{MotionWAM (Ours)}         & \textbf{2.5B} & \textbf{4.9\,Hz} \\
  \bottomrule
\end{tabular}
}}
\caption{\textbf{Deployment efficiency with a single NVIDIA~A100.} 
}
\label{tab:speed}
\vspace{-1.2\baselineskip}
\end{wraptable}
The reported frequency is the \emph{chunk-wise} rate at which the policy emits a whole action chunk.

As shown in Table~\ref{tab:speed}, MotionWAM remains competitive with VLA baselines at matched scale. Crucially, against Cosmos Policy~\cite{cosmospolicy}---another world-model-based policy with a comparable parameter count---MotionWAM is \emph{seven times faster} ($4.9$\,Hz vs.\ $0.7$\,Hz), because Cosmos Policy must iteratively denoise the future video before producing actions whereas MotionWAM reads off intermediate denoising features in a single pass. This confirms that MotionWAM delivers the accuracy gains in Fig.~\ref{fig:comparison} while still meeting the real-time control rates required for closed-loop humanoid balance.

\section{Conclusion}
\label{sec:conclusion}

We present \textbf{MotionWAM}, a real-time World Action Model for autonomous humanoid loco-manipulation from a single egocentric camera. By coupling a Video DiT and a Motion DiT through intermediate denoising features and predicting whole-body motion tokens in a unified action space, MotionWAM, trained with a three-stage egocentric-to-embodiment recipe, outperforms the strongest VLA baseline by over $30\%$ on nine real-world tasks and points to a viable path from large-scale video pretraining to human-like whole-body humanoid policies.

\section{Limitations}
MotionWAM has the following limitations.
\textbf{First}, the Stage~3 fine-tune has been validated only on the Unitree~G1 embodiment; the three-stage paradigm has not been verified on other humanoid platforms to confirm that the recipe transfers across hardware. \textbf{Second}, our evaluation does not include a controlled novel-object generalization study---training and test object sets share visual similarity, and we do not report success on strictly out-of-distribution objects. \textbf{Failure mode:} Relying on a single egocentric camera, MotionWAM falters when the manipulated object leaves the field of view or the head-camera viewpoint drifts from the training distribution, losing visual grounding and stalling; more failure cases are provided in Appendix~\ref{appendix:failure_cases}. 

\bibliography{example}

\appendix
\clearpage

\section{Real-World Task Suite}
\label{appendix:real_tasks}

\subsection{Per-Task Language Prompts}
\label{appendix:real_prompt}

Table~\ref{tab:task_prompts} lists the natural-language task prompts for the real-world task suite listed in Fig.~\ref{fig:exp}.

\begin{table}[htbp]
  \centering
  \setlength{\tabcolsep}{4pt}
  \small
  \begin{tabular}{ll}
    \toprule
    Task ID & Language Prompt \\
    \midrule
    PnP Bottle      & \emph{Pick the bottle and place it in the basket.} \\
    Kick Soccer       & \emph{Kick the soccer into the goal net.} \\
    Retrieve Item        & \emph{Put the bag on the table and then close the drawer.} \\
    Load Cart       & \emph{Push the cart forward and put the clothes on the table into the cart.} \\
    Toss Garbage    & \emph{Throw the garbage into the trash can.} \\
    Lift Basket     & \emph{Take out the clothes basket under the table and place it on the table.} \\
    Stock Shelves   & \emph{Place the drinks on the upper shelf and the vegetables on the lower shelf.} \\
    Wipe Board      & \emph{Clean the whiteboard thoroughly.} \\
    Do Laundry    & \emph{Throw the clothes into the washing machine.} \\
    \bottomrule
  \end{tabular}
  
  \vspace{6pt}
  
  \caption{\textbf{Per-task language prompts for the nine real-world whole-body loco-manipulation tasks.}}
  \label{tab:task_prompts}
  \vspace{-15pt}
\end{table}

\section{Whole-Body Teleoperation Setup}
\label{appendix:teleop}

\subsection{Hardware}
Our real-world experimental platform is built around the Unitree~G1
humanoid robot. The robot is equipped with a pair of ALOHA~2~\cite{aldaco2024aloha} grippers
mounted on its 7-DoF dual arms, providing a 16-DoF upper-body manipulation
interface. Egocentric observations are
captured by a single Intel RealSense D435i RGB camera mounted at the head.
A workstation with a single NVIDIA RTX~4090 GPU runs the policy server,
which the on-board controller queries in closed loop over WebSocket.

\subsection{VR-Based Teleoperation Pipeline}
We collect Stage~3 demonstrations through a VR-based teleoperation pipeline
that produces \emph{whole-body} trajectories
covering locomotion, torso motion, height regulation, foot interaction, and
bimanual hand manipulation in a single action space. The operator wears a
PICO~VR headset and wraps two PICO trackers around the ankles and holds two hand controllers; the
XRoboToolkit~\cite{zhao2025xrobotoolkit} framework streams a synchronized tracking signal which is converted into a full 24-joint
SMPL whole-body pose. SONIC retargets this SMPL pose into 29-DoF joint angles that drive the G1 in closed loop, while the data
exporter records the resulting state, command, and visual streams as a
LeRobot-format episode at 50~Hz, together with the language goal.

\section{Model and Training Configuration}
\label{appendix:model_train}

Table~\ref{tab:model_train} summarizes the model and optimizer hyperparameters of MotionWAM.
The Video DiT backbone is initialised from
Cosmos-Predict2.5-2B~\cite{cosmos25}, with text and VAE encoders frozen
throughout all three stages. Our MotionDiT follows the DiT-B configuration
introduced in DiT4DiT~\cite{ma2026dit4dit}, with the action and state
dimensions enlarged to accommodate our cross-embodiment Stage~2 mixture
(see Appendix~\ref{appendix:data_compose}). All three stages share a single
flow-matching objective with the per-stage modifications described in Sec.~\ref{sec:training}.

\begin{table}[h]
  \centering
  \caption{\textbf{MotionWAM model and training configurations.} A single
  flow-matching objective is reused across all three stages; only the data
  mixture, learning rates, and number of trainable modules change.}
  \label{tab:model_train}
  \setlength{\tabcolsep}{4pt}
  \small
  \begin{tabular}{lccc}
    \toprule
    Parameter & Stage~1 & Stage~2 & Stage~3 \\
    \midrule
    \multicolumn{4}{l}{\emph{Video DiT}} \\
    \quad Base VGM & \multicolumn{3}{c}{Cosmos-Predict2.5-2B~\cite{cosmos25}} \\
    \quad Attention implementation & \multicolumn{3}{c}{flash-attention 2} \\
    \quad Hidden feature dim & \multicolumn{3}{c}{2048} \\
    \quad Conditional frame timestep & \multicolumn{3}{c}{$1\times10^{-4}$} \\
    \quad Future flow inference steps & \multicolumn{3}{c}{1}\\
    \midrule
    \multicolumn{4}{l}{\emph{Motion DiT}} \\
    \quad Action model type      & \multicolumn{3}{c}{DiT-B} \\
    \quad Hidden size / Output dim            & \multicolumn{3}{c}{2560} \\
    \quad Positional embedding  & \multicolumn{3}{c}{interleaved self-attention} \\
    \quad Max sequence length    & \multicolumn{3}{c}{1024} \\
    \quad Action dim ($D_a$)     & --     & 66    & 66 \\
    \quad State dim ($D_s$)      & --     & 64    & 64 \\
    \quad Past action window     & --     & 0     & 0 \\
    \quad Cross attention dim    & \multicolumn{3}{c}{2048} \\
    \quad Dropout / Final dropout & \multicolumn{3}{c}{$0.2$ / True} \\
    \quad Norm type & \multicolumn{3}{c}{AdaLN} \\
    \quad Num inference timesteps & --    & 4     & 4 \\
    \quad Repeated diffusion steps (train) & -- & 8 & 4 \\
    \quad Noise $\beta$ ($\alpha,\beta,s$) & \multicolumn{3}{c}{$(1.5,\,1.0,\,0.999)$} \\
    \quad Num timestep buckets & \multicolumn{3}{c}{$1000$} \\
    \midrule
    \multicolumn{4}{l}{\emph{Training}} \\
    \quad Frozen modules & \multicolumn{3}{c}{VAE + text encoder} \\
    \quad Per-device batch size & 8 & 8 & 8 \\
    \quad Number of GPUs & 128 & 32 & 8 \\
    \quad Max training steps & 100{,}000 & 50{,}000 & 15{,}000 \\
    \quad Num warmup steps & 5{,}000 & 2{,}500 & 750 \\
    \quad Video DiT LR & $1\!\times\!10^{-5}$ & $1\!\times\!10^{-5}$ & $1\!\times\!10^{-5}$ \\
    \quad Motion DiT LR & --    & $1\!\times\!10^{-4}$ & $1\!\times\!10^{-4}$ \\
    \quad LR scheduler & \multicolumn{3}{c}{cosine with min lr} \\
    \quad Min LR & \multicolumn{3}{c}{$5\!\times\!10^{-7}$} \\
    \quad Gradient clipping & \multicolumn{3}{c}{$1.0$} \\
    \quad Gradient accumulation steps & \multicolumn{3}{c}{$1$} \\
    \quad Optimizer & \multicolumn{3}{c}{AdamW}, $(\beta_1,\beta_2){=}(0.9,0.95)$, $\epsilon{=}10^{-8}$ \\
    \quad Weight decay & \multicolumn{3}{c}{$10^{-8}$} \\
    \bottomrule
  \end{tabular}
\end{table}

\section{Baseline Training Configurations}
\label{appendix:baseline_train}

Every baseline is finetuned on the same Stage~3 demonstrations
as MotionWAM (cf.\ Appendix~\ref{appendix:data_compose}) and consumes the
same egocentric RGB observations, language goals, and proprioceptive
states. \textbf{Diffusion Policy (DP)}~\cite{chi2023diffusion} is a representative non-VLA visuomotor policy that models action sequences via a conditional denoising diffusion process; \textbf{ACT}~\cite{zhao2023learning} is a transformer-based action-chunking policy that predicts a short sequence of future actions from visual and proprioceptive inputs in a single forward pass; {$\mathbf{\pi_{0.5}}$}~\cite{intelligence2025pi_} is a generalist VLA policy that couples a pretrained vision--language backbone with a flow-matching action expert and is co-trained on heterogeneous robot data to support open-world manipulation; and \textbf{GR00T-N1.7}~\cite{bjorck2025gr00t} is NVIDIA's open humanoid foundation model, pairing a vision--language reasoning module with a diffusion-transformer action head and pretrained on a large mixture of human and humanoid data for whole-body control. Below we detail the model adaptations and training recipes used for DP, ACT, and our parameter-matched Qwen3DiT baseline; the remaining VLA baselines (GR00T-N1.7 and $\pi_{0.5}$) follow their respective official fine-tuning recipes.

\paragraph{Diffusion Policy (DP)~\cite{chi2023diffusion}.}
We follow the original recipe: a pre-trained ResNet-18~\cite{he2016deep} is
used as the visual encoder and a UNet-based denoiser predicts the action
sequence conditioned on the visual and proprioceptive features. We set the
learning rate to $1\!\times\!10^{-4}$ and the global batch size to $64$, and
train for 40{,}000 steps on 8 NVIDIA A100 GPUs. At inference, we run 100 iterative denoising steps to
progressively transform random noise into actionable whole-body
trajectories. We observe that DP fails on most loco-manipulation tasks, suggesting that the UNet-based DP model has insufficient visual capacity for the egocentric, whole-body setting.

\paragraph{Action Chunking with Transformers (ACT)~\cite{zhao2023learning}.}
To apply the ACT framework to whole-body humanoid loco-manipulation, we modify the action head to generate a 66-dimensional action vector for the Unitree~G1, which includes the two gripper states. Consistent with the public ACT release, we utilize a chunk size of 100 and configure the transformer with 4 encoder layers and 1 decoder layer. The remaining hyperparameters—specifically the learning rate, batch size, and total training steps—remain identical to the DP setup described above.

\paragraph{Qwen3DiT.}
Qwen3DiT is a custom, parameter-matched baseline that we construct as a
direct ablation of the video-pretrained backbone in MotionWAM: it
replaces the Cosmos-Predict2.5 Video DiT with the Qwen3-VL~\cite{bai2025qwen3}
2B foundation model as the visual--language backbone, while keeping the
Motion DiT, the unified motion-latent action space, and the
observation/proprioception interface identical to MotionWAM. 
To provide a
stringent ablation, Qwen3DiT is subjected to the exact same Stage~2
cross-embodiment action post-training and Stage~3 Unitree~G1 fine-tuning
pipeline as MotionWAM (see Appendix~\ref{appendix:data_compose}); only
Stage~1 egocentric video pretraining is omitted, since Qwen3-VL already
provides a static image--text prior that replaces the role of Stage~1 in
MotionWAM. All training settings follow the Stage~2 and Stage~3
columns of Table~\ref{tab:model_train}. This setting isolates the contribution of the video
world model prior over a static VLM prior at matched capacity and under a
matched training budget.

\section{Per-Stage Data Composition}
\label{appendix:data_compose}

The three training stages ingest progressively narrower but progressively
more action-grounded data. Stages~1 and~2 each draw from a weighted mixture
of public datasets while Stage~3 fine-tunes on our in-house Unitree~G1
demonstrations.

\paragraph{Stage 1 — egocentric video pretraining.}
Stage~1 trains the Video DiT alone on a weighted mixture of egocentric
human and humanoid-robot video; the action-related sources are read for
their video streams only and any action labels are ignored. The mixture
weights---budgeted across three domains (human at 30\%, G1-class
humanoids at 50\%, other real robots at 20\%)---are listed in
Table~\ref{tab:stage1_mix}. Inside each domain the per-source weights are
allocated by $\sqrt{\#\text{episodes}}$ to avoid letting any single dataset
dominate.

\begin{table}[h]
  \centering
  \caption{\textbf{Stage~1 video-pretraining mixture.} Weights are
  normalized so that the listed values sum to one. The ``human'' domain
  receives 30\%, ``G1-class humanoid'' 50\%, and ``other real robots''
  20\% of the budget.}
  \label{tab:stage1_mix}
  \setlength{\tabcolsep}{4pt}
  \small
  \begin{tabular}{lll c}
    \toprule
    Source & Embodiment & Weight \\
    \midrule
    EgoDex~\cite{hoque2025egodex}            & Human (egocentric) & 0.300 \\
    GR00T-X-Embodiment-Sim (GR1)~\cite{bjorck2025gr00t} & Fourier GR1 (sim) & 0.255 \\
    RoboCOIN (G1edu/Galbot/Leju)~\cite{wu2025robocoin} & G1edu/Galbot/Leju & 0.080 \\
    GR00T-Teleop-GR1-Robot~\cite{bjorck2025gr00t} & Fourier GR1 (real) & 0.071 \\
    Humanoid-Everyday~\cite{zhao2025humanoideverydaycomprehensiverobotic} & Unitree G1 & 0.047 \\
    UnifoLM-WBT~\cite{unitree2026unifolmwbt} & Unitree G1 (WBT) & 0.023 \\
    PSI-Real~\cite{wei2026psi_0} & Unitree G1 & 0.013 \\
    PSI-Simple~\cite{wei2026psi_0} & Unitree G1 & 0.011 \\
    RoboCOIN (R1\_Lite + RMC-AIDA-L)~\cite{wu2025robocoin} & R1\_Lite, AIDA-L  & 0.200 \\
    \bottomrule
  \end{tabular}
\end{table}

\paragraph{Stage 2 — cross-embodiment action post-training.}
Stage~2 attaches the Motion DiT and co-trains the Video DiT and Motion DiT
on a heterogeneous mixture of action-labelled humanoid datasets, routed
through per-embodiment input/output projectors around the shared Motion
DiT trunk. Action vectors are right-padded to the
maximum action dimension (66) with an accompanying mask that marks the
valid channels for each embodiment, so heterogeneous action layouts can
share the same DiT trunk.

\paragraph{Stage 3 — Unitree~G1 whole-body fine-tuning.}
Stage~3 fine-tunes the full network end-to-end on our in-house
teleoperated whole-body dataset. We collect $200$ episodes per task on the
nine real-world loco-manipulation tasks listed in
Table~\ref{tab:task_prompts}, recorded at 50~Hz on the Unitree~G1 platform described in
Appendix~\ref{appendix:teleop}. All Stage~3 episodes share a single
embodiment tag.

\section{Failure Cases}
\label{appendix:failure_cases}

\begin{figure}[h]
  \centering
  \includegraphics[width=0.5\textwidth]{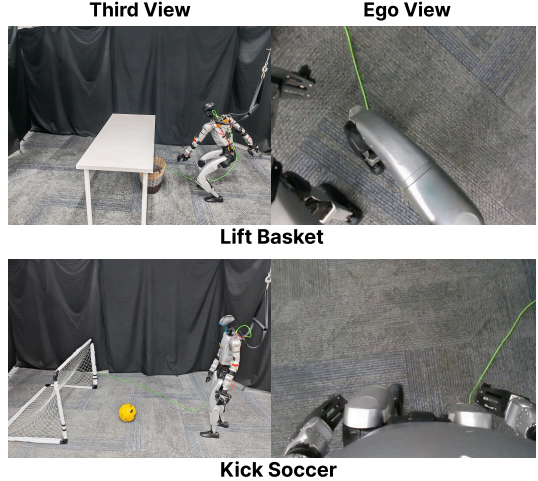}
  \caption{\textbf{Representative failure cases of MotionWAM on the
  nine real-world whole-body loco-manipulation tasks.} Most failures
  trace back to loss of visual grounding when the manipulated object
  exits the egocentric field of view or the head-camera viewpoint
  drifts from the training distribution.}
  \label{fig:failure_cases}
\end{figure}

Figure~\ref{fig:failure_cases} illustrates representative failure modes of
MotionWAM observed across the nine real-world loco-manipulation tasks.
Because MotionWAM relies on a single egocentric head-mounted camera, the
dominant failure mode arises when the manipulated object leaves the
camera's field of view or the head-camera viewpoint drifts away from
the training distribution: visual grounding is lost and the policy
either stalls or commits to an inaccurate whole-body trajectory.

\begin{figure}[h]
  \centering
  \vspace{-1cm}
  \includegraphics[width=1.0\textwidth]{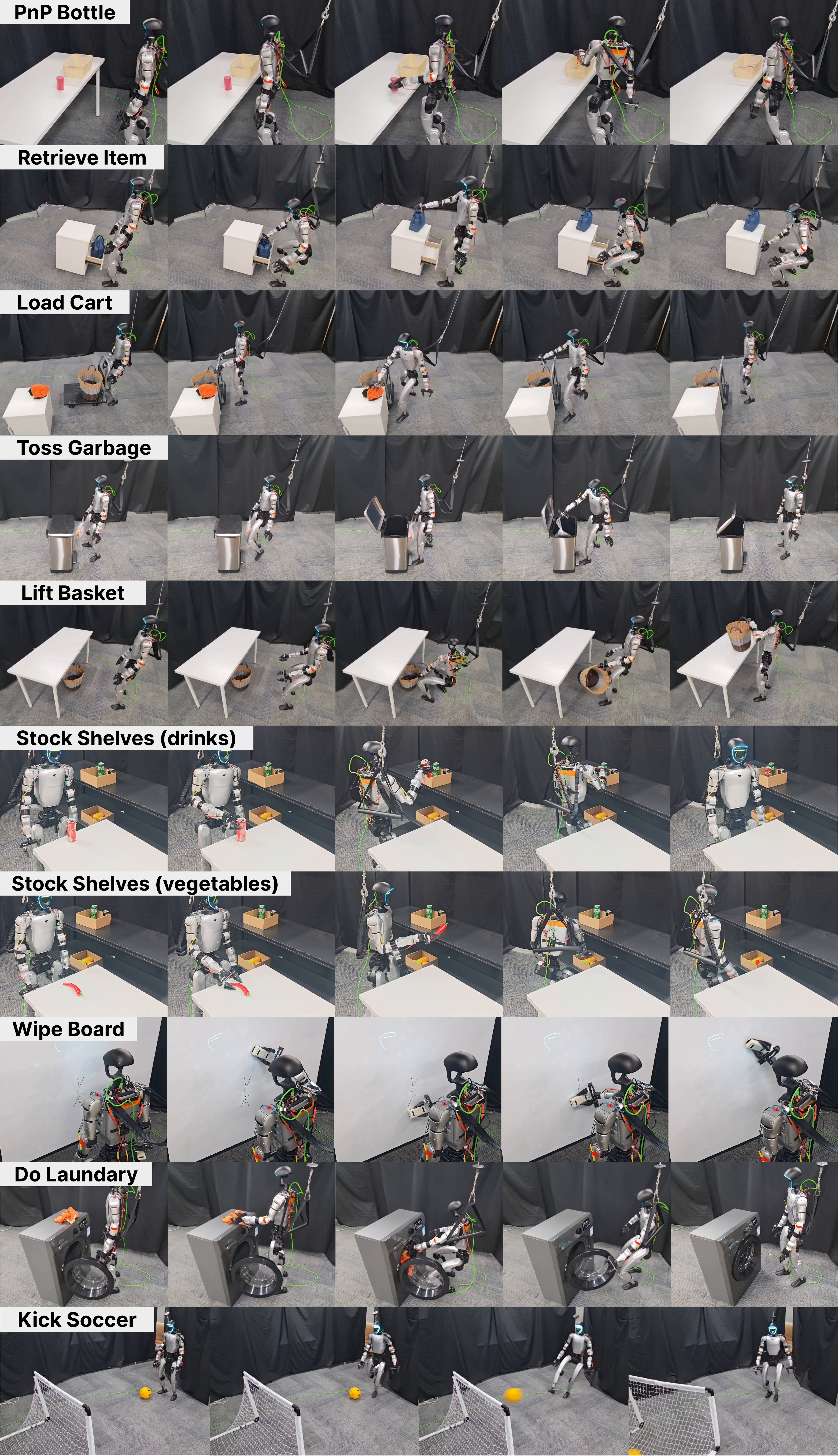}
  \caption{\textbf{Representative MotionWAM inference demonstrations on the nine real-world
  whole-body loco-manipulation tasks.}}
  \label{fig:full_demos}
\end{figure}

\end{document}